\documentclass[conference]{IEEEtran}
\IEEEoverridecommandlockouts
\usepackage{amsmath,amssymb,amsfonts}
\usepackage{algorithmic}
\usepackage{graphicx}

\usepackage{textcomp}
\usepackage{xcolor}
\usepackage{cleveref}

\usepackage{float}
\usepackage[numbers]{natbib}
\usepackage{fancyhdr}
\usepackage{url}
\def\BibTeX{{\rm B\kern-.05em{\sc i\kern-.025em b}\kern-.08em
\usepackage{natbib}
    T\kern-.1667em\lower.7ex\hbox{E}\kern-.125emX}}

\begin{document}

\title{
   Improved Sales Forecasting using Trend and Seasonality Decomposition with LightGBM
   }

\author{\IEEEauthorblockN{Tong Zhou}
\IEEEauthorblockA{\textit{Department of Computer Science} \\
\textit{Johns Hopkins University}\\
Baltimore, United States \\
tzhou11@jhu.edu\\ 
\noindent \textit{DOI: } \url{https://doi.org/10.1109/ICAIBD57115.2023.10206380}
}
}

\maketitle
\thispagestyle{fancy}
\lhead{}
\lfoot{}
\cfoot{\small{© © 2023 IEEE. Personal use of this material is permitted. Permission from IEEE must be obtained for all other uses, in any current or future media, including reprinting/republishing this material for advertising or promotional purposes, creating new collective works, for resale or redistribution to servers or lists, or reuse of any copyrighted component of this work in other works.}}
\rfoot{}

\begin{abstract}
   Retail sales forecasting presents a significant challenge for large retailers such as Walmart and Amazon, due to the vast assortment of products, geographical location heterogeneity, seasonality, and external factors including weather, local economic conditions, and geopolitical events. Various methods have been employed to tackle this challenge, including traditional time series models, machine learning models, and neural network mechanisms, but the difficulty persists. Categorizing data into relevant groups has been shown to improve sales forecast accuracy as time series from different categories may exhibit distinct patterns. In this paper, we propose a new measure to indicate the unique impacts of the trend and seasonality components on a time series and suggest grouping time series based on this measure. We apply this approach to Walmart sales data from 01/29/2011 to 05/22/2016 and generate sales forecasts from 05/23/2016 to 06/19/2016. Our experiments show that the proposed strategy can achieve improved accuracy. Furthermore, we present a robust pipeline for conducting retail sales forecasting.
\end{abstract}
\section{Introduction}
Sales or demand forecasting is essential for businesses operating across different regions, states or countries, such as Walmart and Amazon. Reliable demand forecasting enables them to make wise decisions regarding their inventory, pricing and marketing tactics. By accurately predicting future demand, businesses can also optimize inventory levels, reduce wastes and increase profitability. Correct demand forecasting is also crucial for a country's stable supply chain (\cite{hyndman2018forecasting,trapero2015identification}). From the consumers side, a stable supply chain ensures availabilities of goods and services. This helps to maintain consumer confidence and prevent panic buying. For the businesses side, a stable supply chain allows them to operate smoothly and efficiently, as the risks of disruptions and downtime can be minimized.

The application of both traditional time series models and modern machine learning and AI techniques for sales and demand forecasting has garnered significant attention in the field of forecasting. Traditional time series models, such as ARIMA, SARIMA, and Exponential Smoothing State Space Model (ETS), have been widely used for their simplicity, interpretability, and ability to capture linear trends and seasonality. These models have demonstrated effectiveness in various sales and demand forecasting scenarios, making them a popular choice for many practitioners.

However, with the increasing availability of large datasets and the growing complexity of retail sales data, modern machine learning and AI techniques have emerged as powerful alternatives. Methods such as artificial neural networks (ANNs), support vector machines (SVMs), random forests, and gradient boosting machines (GBMs) have shown promise in handling non-linear relationships, high-dimensional data, and complex interactions between variables. Moreover, recent advancements in deep learning, such as recurrent neural networks (RNNs) and long short-term memory (LSTM) networks, have further improved the accuracy and adaptability of sales and demand forecasts by capturing long-term dependencies in time series data.

Combining traditional time series models with machine learning and AI techniques can lead to a more comprehensive forecasting approach. Hybrid models, which integrate the strengths of both methodologies, have been proposed as a means to enhance forecast accuracy and robustness. Ensemble methods, which combine the predictions of multiple models, have also been explored to capitalize on the diverse capabilities of different forecasting techniques.

This paper presents such an hybrid model.We harness the power of the LightGBM algorithm, a high-performance gradient boosting framework, to generate sales forecasts for each time series under the assumption of stationarity. By assuming stationarity, we are able to  utilize the LightGBM's ability to capture complex non-linear relationships and interactions between variables. We employ the Prophet model, a robust and flexible forecasting tool developed by Meta, to achieve more precise sales forecasts across a variety of aggregated levels within the time series data. The Prophet model is particularly adept at handling the irregularities and seasonality commonly found in sales data, making it a valuable complement to the LightGBM approach. Finally, we combine the results of these two powerful methods, creating an hybrid model that leverages their respective strengths to generate robust and accurate sales forecasts for each time series. This fusion of state-of-the-art techniques not only enhances the overall forecasting performance but also makes the approach more resilient to diverse and challenging retail scenarios. By integrating the LightGBM and Prophet models, this paper presents an intriguing and effective solution to the complex task of sales forecasting, offering valuable insights and a practical tool for researchers and practitioners alike. 

\section{related work}
The history of sales forecasting is long and complex, with many different methods and models proposed over the years. In recent years, machine learning and deep learning techniques have gained popularity for their ability to handle large and complex data sets. This section provides an overview of some of the key papers and works related to sales forecasting.

In the early days of sales forecasting, statistical models such as ARIMA (autoregressive integrated moving average) and exponential smoothing were popular. \cite{makridakis1983averages} conducted the M3 competition, which aimed to compare the accuracy of different forecasting methods. The results showed that neural networks outperformed statistical models in some cases, sparking interest in machine learning methods for forecasting.

\cite{fiorucci2016time} surveys the Theta method, which combines ARIMA and exponential smoothing. This method has been shown to be highly accurate for short-term forecasting. \cite{armstrong2001extrapolation} proposed the extrapolation method, which involves extending a time series trend line to make future predictions. This method is simple but can be prone to errors if the underlying trend changes.

In recent years, deep learning models such as recurrent neural networks (RNNs) and long short-term memory (LSTM) networks have gained popularity for their ability to capture complex patterns in time series data. \cite{shao2015rolling}  proposed the use of a deep belief network for sales forecasting, which achieved high accuracy on a large data set. A survey on sales forecast using deep learning can be found in \cite{ensafi2022time}.

The history of sales forecasting is rich with different methods and models proposed over the years. While statistical models such as ARIMA and exponential smoothing are still widely used, machine learning and deep learning techniques have become increasingly popular in recent years due to their ability to handle large and complex data sets.

\section{data}
The Walmart's sales dataset is a large and complex time series dataset that includes daily sales data for thousands of products across ten stores of Walmart in California (CA), Texas (TX) and Wisconsin (WI). The dataset includes information on various products such as food, household essentials, and hobbies, sold across three categories - food, household, and hobbies. The sales data spans a period of almost five years, from January 1, 2011, to June 30, 2016, with a total of 1969 days of sales data. The original dataset contains $30490$ time series. Based on different levels of aggregation, they can be built into a hierarchical structure of time series with  $42840$ time series. \Cref{tab:data} illustrates such aggregation.

The data also includes information on promotions, price changes, holidays, and other external factors that may impact sales. In addition to the sales data, the dataset includes various supplementary data files such as calendar information, prices of products, and information on special events such as Super Bowl Sundays.

\begin{table}[ht]
\caption{Number of Walmart series per aggregation level.}
\begin{tabular}{lc}
\hline
\multicolumn{1}{c}{\bfseries Aggregation Level} & {\bfseries Number of Series} \\ \hline
  all products, aggregated for all stores/states & 1 \\
 all products, aggregated for each State & 3 \\
 all products, aggregated for each store & 10 \\
 all products, aggregated for each category & 3 \\
 all products, aggregated for each department & 7 \\
 all products, aggregated for each State/category & 9 \\
 all products, aggregated for each State/department & 21 \\
 all products, aggregated for each store/category & 30 \\
 all products, aggregated for each store/department & 70 \\
 product x, aggregated for all stores/states & 3,049 \\
 product x, aggregated for each State & 9,147 \\
 product x, aggregated for each store & 30,490 \\
 \multicolumn{1}{c}{\bfseries{Total}} & 42,840 \\ \hline
\end{tabular}
\label{tab:data}
\end{table}
 
\subsection{Trend}
\begin{figure}
   \centering
   \includegraphics[scale=.45]{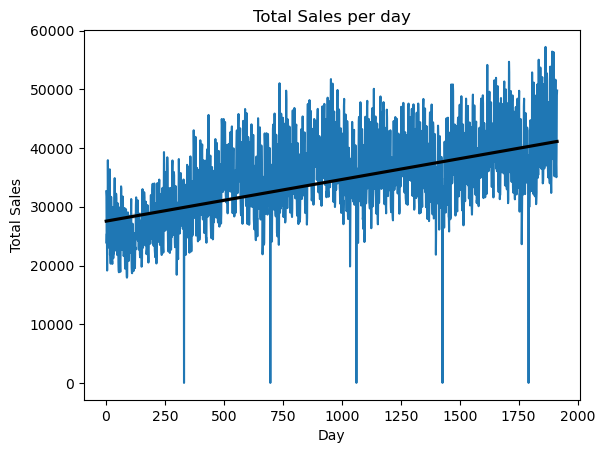}
   \caption{Total sales per day}
   \label{fig:sales}
\end{figure}

We add up all sales together for each day across states, stores and items, and plot the total sales from day 1 to the day 1941 in Fig. \ref{fig:sales}. This figure exhibits a clear upward trending. Whatever time series models we may utilize, this figure indicates that trend cannot be ignored. 

Another message from the figure suggests that decision-tree based machine learning (ML) models may not be appropriate for forecasting the total sales of the 1941 days, as those models are unable to do extrapolation. Extrapolation refers to the ability of a model to make predictions beyond the range of the data used to train the model. Decision-tree based models, such as Random Forest or XGBoost, are tree-based models that partition the feature space into regions based on simple rules. These models can be effective for modeling complex nonlinear relationships and interactions between features, but they may not be well-suited for extrapolation. This is because decision-tree models make predictions based on the average of the target variable within each partition of the feature space. If a new data point falls outside the range of the training data used to construct the partitions, the model will not have any information about how to predict the target variable for that point, and will instead use the average of the nearest partitions. This can lead to inaccurate or unrealistic predictions, especially if the new data point is significantly different from the training data.
\subsection{Heterogeneous trends across stores}
\begin{figure}
   \centering
   \includegraphics[scale=.45]{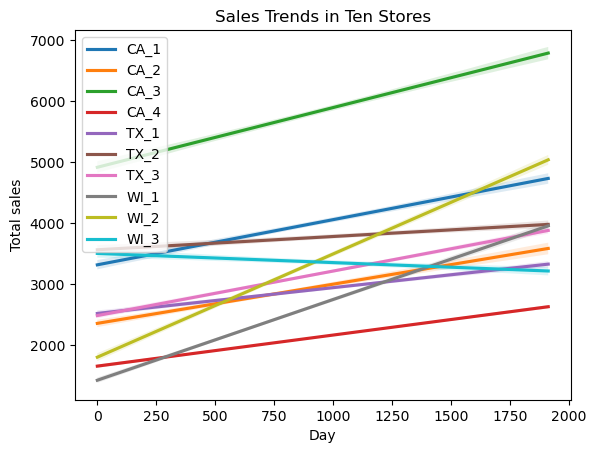}
   \caption{Sales trends in 10 stores}
   \label{fig:trends_store}
\end{figure}

As is stated in the last section, the challenges of conducting demand forecasts is vastly distinct time series patterns if we split the entire dataset into different categories. We first examine the trends of the ten stores in the Walmart dataset in Fig. \ref{fig:trends_store}. The fact that the time series patterns of these ten stores are distinct suggests that using different time series models for each store may be appropriate. This is because different time series models may be better suited to capture the unique patterns of variation in each store's sales data, such as differences in seasonality, trend, and cyclicality.

\begin{figure}
   \centering
   \includegraphics[scale=.45]{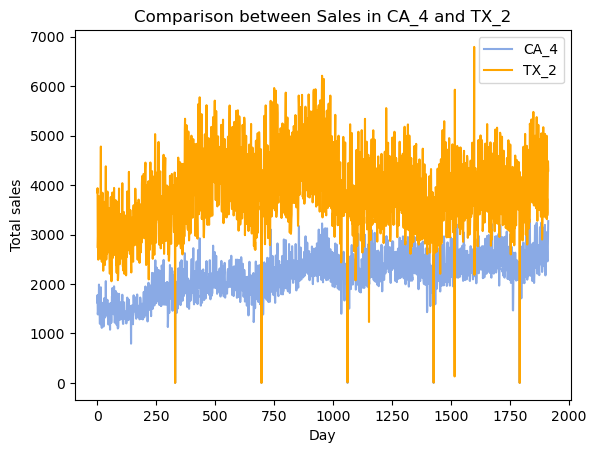}
   \caption{Sales trends in CA$\_$4 and TX$\_$2}
   \label{fig:CA_TX}
\end{figure}

We pick two stores CA$\_$4 and TX$\_$2. Fig. \ref{fig:CA_TX} appears that the time series patterns of the CA$\_$4 and TX$\_$2 stores are quite different, with different levels, seasonality, and sales variances for each day. These differences suggest that using different time series models for forecasting may be necessary.
\subsection{Heterogeneous time series patterns across states}
\begin{figure}
   \centering
   \includegraphics[scale=.45]{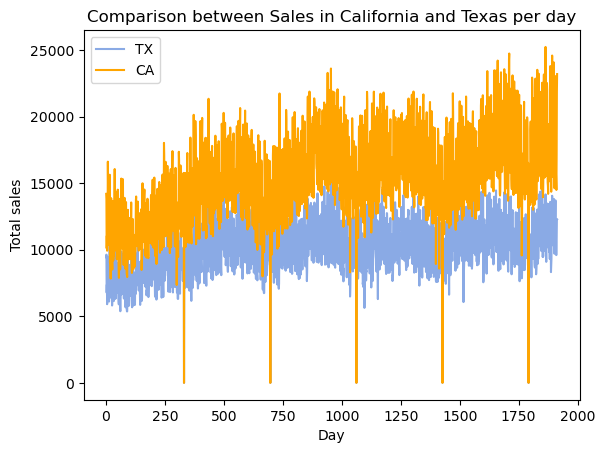}
   \caption{Sales Distribution from $\text{day}\_1$ to $\text{day}\_1913$}
   \label{compare}
\end{figure}

Fig. \ref{compare} displays the total sales of the 1941 days in Texas and California. It is clearly shown that both series have different levels and trends.Therefore, it is inappropriate to model them using the same time series structure. If the same time series model is applied to both series, it may not capture the unique patterns of variation in each series, and the resulting forecasts may be inaccurate. For example, it seems that CA's sales has a relative stronger seasonality pattern by observing a sudden drop every year, a model that assumes the same seasonal pattern for both series may not capture the true seasonality in each series.
\subsection{Sales distribution}
\begin{figure}
   \centering
   \includegraphics[scale=.45]{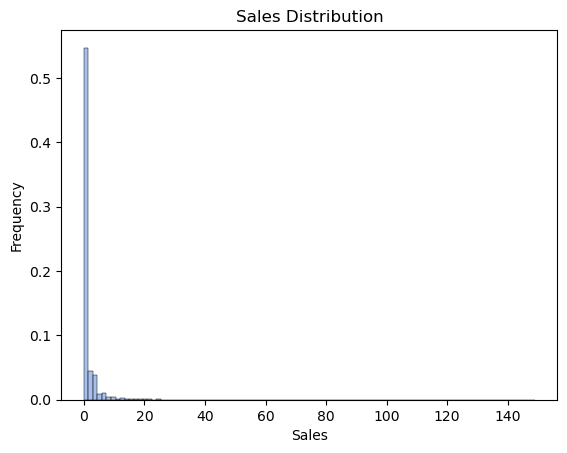}
   \caption{Sales Distribution from $\text{day}\_1$ to $\text{day}\_1913$}
   \label{sales_dist}
\end{figure}

Based on the plot of the labels (sales) distribution in Fig. \ref{sales_dist}, it appears that over $50\%$ of sales are 0, and the distribution is heavily right-skewed. This suggests that the data may not conform to the assumptions of a normal distribution, which is the assumption underlying the use of mean squared loss. Mean squared loss is a commonly used loss function for regression problems, as it penalizes large differences between predicted and actual values. However, when the data has a non-normal distribution, mean squared loss may not be appropriate, as it can lead to biased or inaccurate predictions.

\subsection{Feature Engineering}
\subsubsection{Constructing hierarchical time series}
Based on our data explorations, it appears that the sales data has a hierarchical structure, with sales data at the level of individual products, departments, stores, and states. This suggests that using a hierarchical time series framework may be a good approach to modeling and forecasting the sales data.

A hierarchical time series framework involves modeling the data at multiple levels of aggregation, with different models for each level. This allows us to capture the unique patterns of variation at each level, while also accounting for the dependencies and relationships between the different levels. For example, a model at the product level can capture the unique seasonal patterns and trends of each product, while a model at the store level can capture the effects of local factors such as promotions or regional events. Modeling the sales data at the state level can also be important for capturing the unique patterns of variation within each state, such as differences in consumer preferences, local economic conditions, or regulatory policies. For example, a store in California may have different sales patterns compared to a store in Texas, due to differences in demographic factors, climate, or local competition.

\subsubsection{Converting a wide dataset into a long format}
The size of the dataset in terms of its number of columns, which is 1913, is quite large. As a result, the dataset is considered to be "wide" in nature. However, in order to model the data as a regression problem and use tree-based ML algorithms, it is necessary to convert the dataset into a "long" format.

Converting the dataset to a "long" format involves setting up two columns: one column for sales data in the current year, and another column for the label, which is the next year's sales. This transformation is necessary as it enables the data to be modeled as a regression problem, where the objective is to predict the next year's sales based on the current year's sales and other relevant features.

Additionally, incorporating more time series features, such as the last 7 and 28 days' mean sales, calendar time and lag features,  can help to capture the temporal dependencies and patterns in the data. This can be particularly important in retail sales forecasting, as sales patterns can be heavily influenced by trends and seasonal patterns.

\section{Method}
\begin{figure}[t]
   \centering
   \includegraphics[width=.5\textwidth, height=0.5\textwidth]{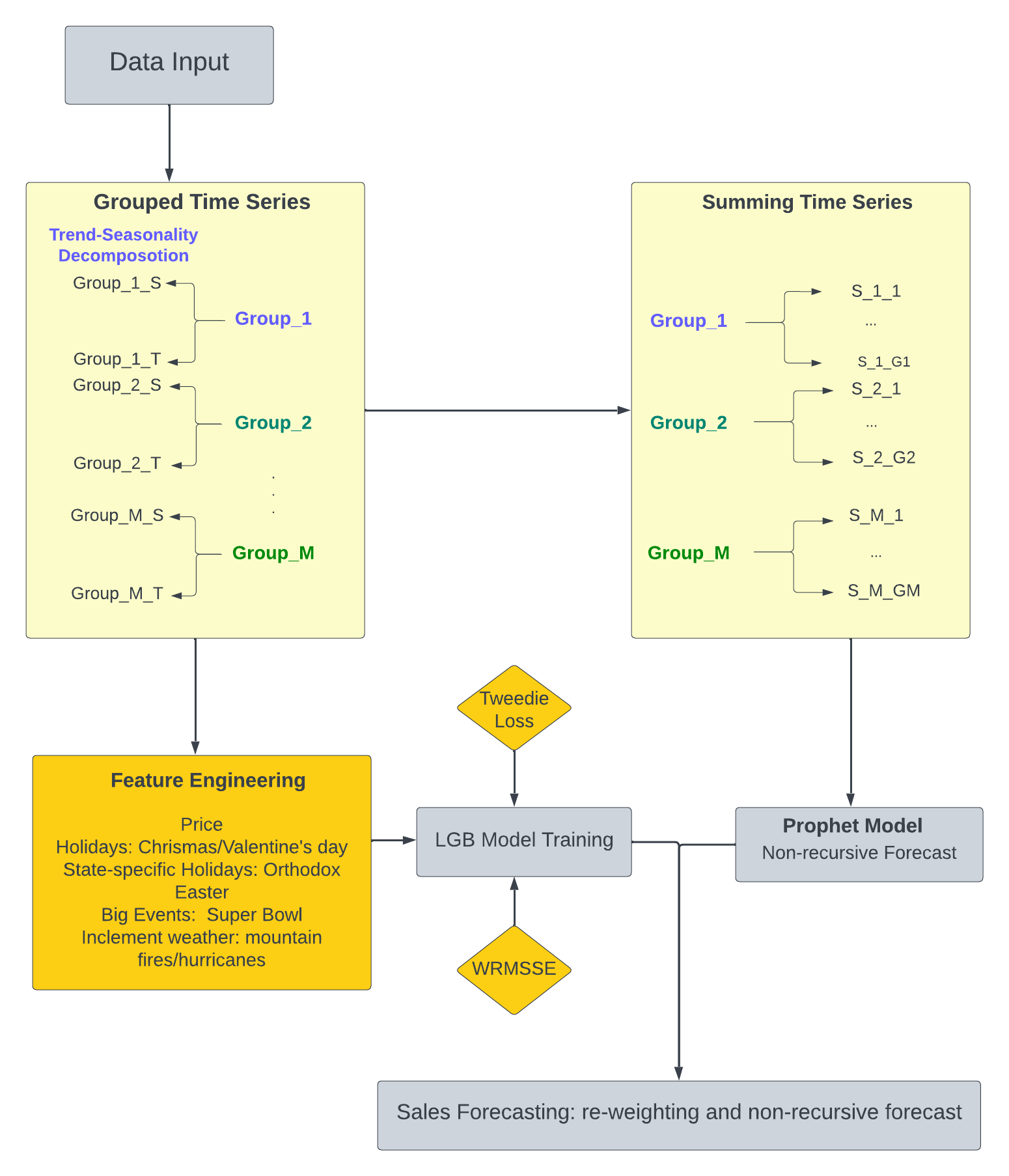}
   \caption{Proposed pipeline for sales forecasting}
   \label{fig:model}
\end{figure}

The overview of our methodology is illustrated in Fig. \ref{fig:model}. The idea is that in the left block, the entire dataset is first be split into different categories based on stores, states, product categories etc.. Then feature engineering is conducted, including adding more features, dataset formatting conversions etc.. Last, LightGBM is used to do forecast on each time series. It is noted that in the whole LightGBM pipeline, we do not make any aggregation on those time series.

\subsection{LightGBM}
LightGBM (Light Gradient Boosting Machine) is an open-source, distributed machine learning library that uses gradient boosting algorithms to train models. It is developed by Microsoft and is known for its high speed and accuracy, making it a popular choice for large-scale and complex datasets.  LightGBM uses a novel technique called Gradient-based One-Side Sampling (GOSS) to speed up the training process. This technique only selects a small fraction of the data points for each iteration, reducing the computational cost significantly. It uses a technique called Exclusive Feature Bundling (EFB) to reduce memory usage. It combines similar features into bundles, reducing the number of features without losing information.

As is discussed above, LightGBM may not perform well in extrapolation. In our sales forecast problem, we have seen many time series exhibit clear upward trend, rendering merely using LightGBM inappropriate. 

This issue can be remedied by combining LightGBM and trend forecasting models. LightGBM is particularly good at handling large datasets with many features and detecting nonlinear relationship between features and labels. On the other hand trend forecasting models, such as Prophet, ARIMA and STL Decomposition, can help capture the seasonality and trend patterns in the data. By combining the strengths of LightGBM and trend forecasting models such as Prophet, it is possible to build a robust and accurate forecasting system by assuming the input time series of the LightGBM is stationary and its output can serve as weights to adjust the total sales forecast by Prophet.

\subsection{Evaluation Metrics}
The Weighted Root Mean Squared Scaled Error (WRMSSE) is used as the evaluation metric because it is specifically designed to evaluate the accuracy of hierarchical time series forecasts. For example, the sales data in the Walmart dataset has a hierarchical structure, with sales data at the level of individual products, departments, stores, and regions. The total number of time series in the original dataset is $30490$. After exhausting all possibilities of merging them by items, categories, states and stores, we end up having a total of $42840$ rows of data. 

The WRMSSE takes this hierarchical structure into account and provides a measure of the accuracy of the forecasts at each level, while also accounting for the dependencies and relationships between the different levels.
Consider the $i$-th time series, its RMSSE is defined as
\begin{align*}
   \text{RMSSE}_{i} = \sqrt{\frac{1}{h} \frac{\sum_{t=T+1}^{T+h}(Y_{it}-\widehat{Y}_{it} )^2}{\frac{1}{T-1}\sum_{t=2}^{T}(Y_{it}-Y_{i,t-1})^2}} 
\end{align*}
and
\begin{align*}
   \text{WRMSSE} = \sum_{i=1}^{N}w_{i}\cdot \text{RMSSE}_{i}
\end{align*}
where $w_{i}$ denotes the weighted total sales from $T - 28$ to  $T$ and  $\sum_{i=1}^{N}w_{i}=1$. In the Walmart dataset, for example, the original dataset contains 39490 time series sequence, but there are in total $42840$ sequences when all hierarchical time series sequences are taken into account.
\subsection{Tweedie-based loss function}
From Fig. \ref{sales_dist}, the sales distribution in the train data exhibits a Poisson-like distribution: non-negative response values, right-skewed and long-tailed distribution. It reminds us of using a different loss function than the mean squared loss that is more appropriate for Gaussian distributions.

Tweedie distribution belongs to the class of exponential dispersion models (EDM). It provides a flexible modeling for non-negative and right-skewed data (\cite{dunn2005series}). Thus, we shall use Tweedie-based loss function for our model. We use the usual trick of the negative log-likelihood as the loss function. Suppose  each sales $S_{t}$ follows a Tweedie distribution, the loss function  can be written as 
\begin{align*}
   \mathsf{Loss}(\mathbf{S}, \widehat{\mathbf{S} }  ) = -\sum_{t=1}^{T}\left(S_{t}  \cdot \frac{\widehat{S}_{t}^{1-p} }{1-p} + \frac{\widehat{S}_{t}^{2-p} }{2-p}\right),
\end{align*}
where $S_{t}$ is the ground truth, $\widehat{S}_{t} $ is the predicted sales, and $p \in (1, 2)$ is a hyperparameter characterizing the power relation between distribution mean and variance. When $p$ is close to  $1$, it becomes close to Poisson distribution, and when  $p$ is close to  $2$, it becomes close to Chi-square distribution.

\subsection{Prophet}
Prophet is a time series forecasting library developed by Facebook that is designed to make it easy for analysts and developers to create accurate forecasts for time series data. It uses a decomposable model that allows it to capture trends, seasonality, and other time-varying effects in a flexible and scalable way. It is particularly well-suited for datasets with strong seasonal effects and long-term trends. It can also handle missing data and outliers, and provides a range of customizable parameters to fine-tune the model's performance.

Prophet decomposes a time series into four components
\begin{align}
   Y_{t} = T(t) + S(t) + H(t) + R(t), \forall t
   \label{eqn:decomposition}
\end{align}
where $T(t), S(t), H(t)$ and  $R(t)$ represent trend, seasonality, holiday effect and residuals, respectively.

\subsection{Trend-seasonality decomposition}
In addition to dividing the entire dataset into several groups based on observed characteristics, we make an additional splitting based on whether the trend is stronger or the seasonality is stronger in one time series. First, we define a concept \textit{score} to measure the influence of trend and seasonality that are  decomposed in \eqref{eqn:decomposition}

The score of trend and seasonality scores $\mathsf{Score}_{T}$ and $\mathsf{Score}_{S}$ are defined as follows
\begin{align*}
   \mathsf{Score}_{\mathsf{T}} &=  \max \left( 0, 1 - \frac{\mathsf{Var}(R_{t})}{\mathsf{Var}(T_{t} + R_{t})}  \right)\\
   \mathsf{Score}_{\mathsf{S}} &=  \max \left( 0, 1 - \frac{\mathsf{Var}(R_{t})}{\mathsf{Var}(S_{t} + R_{t})}  \right).\\
\end{align*}
where $\mathsf{Var}(\cdot) $ denotes the variance of a random variable. The intuition, for instance, is that if $\mathsf{Var}(T_t + R_t)$ is relatively larger, the score of trend is larger, and thus this time series is considered to be a ``trend" type.

Based on which score is greater, all sequences in each group are divided into two groups. Each group will be fit into different LightGBM models. This approach allows different models can discover distinct patterns resulting from different sources.

\subsection{Forecasting strategy}
We present the final step of sales forecasting. Suppose for any group, it has $g$ sequences of time series. The LightGBM predicts that the sales at time  $t$ is  $\widehat{p}_{1t}, \dots, \widehat{p}_{gt}$. Since the LightGBM model assumes all the time series are stationary without trend, a useful way is to think of the predicted values as being probabilities, or relative contributions to the final output. Also, suppose the total predicted sales of this group at $t$ using Prophet is $G_{t}$, then $G_{t}$ can be allocated by the weights obtained from the LightGBM, i.e., the ultimate forecast sales for the time series at $t$ is given by 
\begin{align*}
   \mathsf{Sales}_{j t} = \frac{\widehat{p}_{j t}}{\sum_{j = 1}^{g}\widehat{p}_{j t}} \cdot G_{t}, \forall j = 1, \dots, g. 
\end{align*}

It is also noted that we use a non-recursive fashion to avoid forecasting error propagation. Since we are doing multiple forecasts for consecutive $28$ days, if we adopt a recursive forecast, which means that next predictions depend on previous predictions, then previous forecasting errors will be accumulated and later forecasts will be disastrously inaccurate. 

\section{results and discussions}
To demonstrate our trend-seasonality decomposition strategy achieve improved accuracy and justify the Tweedie-based loss function, we conduct several experiments. Their comparisons are summarized in \Cref{tab:performance}
\begin{table}[ht]
   \centering
   \caption{Performance Comparison of Models}
   \begin{tabular}{ c c c}
      \hline
      Model & Loss Function & WRMSSE  \\ \hline
      Linear Regression & Tweedie & 1.120 \\
      SVM & Tweedie & 1.083 \\ 
      LSTM & Tweedie & 0.802 \\ 
      No T-S Decomposition & MSE & 0.963 \\ 
      No T-S Decomposition & Tweedie & 0.832 \\ 
      T-S Decomposition & MSE & 0.767 \\ 
      \textbf{T-S Decomposition} & \textbf{Tweedie} & \textbf{0.614} \\ \hline
   \end{tabular}
   \label{tab:performance}
\end{table}

\Cref{tab:performance} indicate that the trend-seasonality (T-S) decomposition strategy can improve the accuracy of our models. The table compares the performance of models with and without T-S decomposition using different loss functions. The results show that the models with T-S decomposition outperformed those without, regardless of the loss function used. Additionally, the Tweedie-based loss function consistently outperformed the mean squared error (MSE) loss function in both types of models. Specifically, the model with T-S decomposition and Tweedie loss function achieved the lowest WRMSSE (weighted root mean squared scaled error) score of 0.614, demonstrating the effectiveness of our approach. Also, our proposed model beats linear regression, SVM and LSTM in terms of  WRMSSE.
\section{Conclusion}
Sales forecasting poses significant challenges for large corporations such as Walmart, primarily due to the complex and multifaceted nature of the retail industry. This paper contributes to the growing filed of using AI/ML in sales forecasting. 

First, we introduce a robust and scalable framework for conducting sales forecasts, designed to accommodate the intricate demands of large corporations. This framework integrates the LightGBM  and traditional time series decomposition techniques, allowing for the efficient handling of vast product assortments and diverse store locations. By leveraging machine learning techniques and adaptive time series models, our framework can effectively capture complex demand patterns and accommodate fluctuations due to strong trend and seasonality. Moreover, the scalable nature of the framework enables it to process large volumes of data, ensuring rapid and accurate sales predictions even as the company grows or market conditions change. With its capacity for continuous improvement through iterative learning, this framework delivers a powerful and adaptable solution for sales forecasting, ultimately supporting informed decision-making and strategic planning in the dynamic retail landscape.

This paper further posits that utilizing trend-seasonality decomposition to group time series data is a highly effective method for enhancing sales forecasts. By using the proposed scores for trend and seasonality, we can identify and separate the long-term growth patterns from the cyclical fluctuations caused by seasonal factors. This separation allows for a more focused analysis of the individual components, enabling the development of tailored forecasting models that account for the distinct characteristics of each group.

This paper also suggests that employing a Tweedie-based loss function is a valid approach to enhance the accuracy of sales forecasts. The Tweedie distribution, a member of the exponential dispersion family, is particularly well-suited for modeling non-negative, discrete or continuous data, often exhibiting a preponderance of zeros, such as retail sales. By leveraging the Tweedie distribution, we can account for both the zero-inflated nature of sales data and the high variability of sales figures. By incorporating the Tweedie-based loss function into our robust and scalable framework, we can ensure that our sales forecasts are not only adaptable to a wide variety of retail scenarios but are also capable of effectively handling the unique challenges posed by zero-inflated and highly variable sales data.
\section*{Acknowledgment}
We are especially grateful to the organizers of the M5 sales forecasting Kaggle competition for providing the dataset that was instrumental in conducting this research. 
\bibliographystyle{IEEEtranN}
\bibliography{reference}

\end{document}